\documentclass{article}

\usepackage{microtype}
\usepackage{graphicx}
\usepackage{subfigure}
\usepackage{booktabs} % for professional tables

\usepackage{hyperref}

\usepackage[accepted]{icml2024}

\usepackage{amsmath}
\usepackage{amssymb}
\usepackage{mathtools}
\usepackage{amsthm}

\usepackage[capitalize,noabbrev]{cleveref}

\usepackage{enumitem}
\usepackage{amsmath,amsfonts,bm}

\def\eqref#1{equation~\ref{#1}}
\def\1{\bm{1}}

\def\vtheta{{\bm{\theta}}}

\def\vx{{\bm{x}}}
\def\vy{{\bm{y}}}
\def\vz{{\bm{z}}}

\DeclareMathAlphabet{\mathsfit}{\encodingdefault}{\sfdefault}{m}{sl}
\SetMathAlphabet{\mathsfit}{bold}{\encodingdefault}{\sfdefault}{bx}{n}

\newcommand{\E}{\mathbb{E}}

\DeclareMathOperator*{\argmax}{arg\,max}

\newcommand{\eg}{\textit{e.g.}}
\newcommand{\ie}{\textit{i.e.}}

\usepackage[capitalize]{cleveref}
\crefname{section}{Sec.}{Secs.}
\Crefname{section}{Section}{Sections}
\Crefname{table}{Table}{Tables}
\crefname{table}{Tab.}{Tabs.}
\crefformat{section}{\S#2#1#3} % see manual of cleveref, section 8.2.1
\crefformat{subsection}{\S#2#1#3}
\crefformat{subsubsection}{\S#2#1#3}

\usepackage{tikz}
\usetikzlibrary{positioning, shapes.geometric, arrows.meta, calc}

\theoremstyle{plain}

\theoremstyle{definition}

\theoremstyle{remark}

\usepackage[textsize=tiny]{todonotes}

\icmltitlerunning{Co-Supervised Learning}

\begin{document}

\twocolumn[
\icmltitle{Co-Supervised Learning:\\
Improving Weak-to-Strong Generalization with Hierarchical Mixture of Experts}



\icmlsetsymbol{equal}{*}

\begin{icmlauthorlist}
\icmlauthor{Yuejiang Liu}{epfl}
\icmlauthor{Alexandre Alahi}{epfl}
% \icmlauthor{Firstname1 Lastname1}{yyy}
\end{icmlauthorlist}

\icmlaffiliation{epfl}{École Polytechnique Fédérale de Lausanne (EPFL)}

\icmlcorrespondingauthor{}{yuejiang.liu@epfl.ch/stanford.edu}

\icmlkeywords{Machine Learning, ICML}

\vskip 0.3in
]



\printAffiliationsAndNotice{}  % leave blank if no need to mention equal contribution

\begin{abstract}
  Steering the behavior of a strong model pre-trained on internet-scale data can be difficult due to the scarcity of competent supervisors. Recent studies reveal that, despite supervisory noises, a strong student model may surpass its weak teacher when fine-tuned on specific objectives. Yet, the effectiveness of such weak-to-strong generalization remains limited, especially in the presence of large capability gaps. In this paper, we propose to address this challenge by harnessing a diverse set of specialized teachers, instead of a single generalist one, that collectively supervises the strong student. Our approach resembles the classical hierarchical mixture of experts, with two components tailored for co-supervision: (i) we progressively alternate student training and teacher assignment, leveraging the growth of the strong student to identify plausible supervisions; (ii) we conservatively enforce teacher-student and local-global consistency, leveraging their dependencies to reject potential annotation noises.
We validate the proposed method through visual recognition tasks on the OpenAI weak-to-strong benchmark and additional multi-domain datasets. Our code is available at \url{https://github.com/yuejiangliu/csl}.

\end{abstract}

% Begin a group for the quote with Palatino font and right alignment
{
% \fontfamily{ppl}\selectfont
% \fontfamily{ptm}\selectfont
% \fontfamily{ebgaramond}\selectfont
% \fontfamily{pbk}\selectfont
% \fontfamily{bch}\selectfont
% \fontfamily{Baskervaldx}\selectfont
{
\it
“When three are walking together, I am sure to find teachers among them.
I will select their good qualities and follow them, their bad qualities and avoid them.”
\vspace{-6pt}
\begin{flushright}
--- The Analects of Confucius
\end{flushright}
\vspace{6pt}
}

\section{Introduction}
\label{sec:intro}
Recent years have witnessed enormous advances of large neural networks in the presence of two elements: massive unlabeled data for pre-training and high-quality labeled data for fine-tuning. The former, such as texts~\citep{brownLanguageModelsAre2020,touvronLlamaOpenFoundation2023}, images~\cite{rameshZeroShotTexttoImageGeneration2021a,rombachHighResolutionImageSynthesis2022}, and videos~\citep{videoworldsimulators2024} sourced from the internet, empowers large neural networks to acquire a broad spectrum of world knowledge in an unsupervised or self-supervised manner. The latter, manifesting as demonstrations~\citep{brownLanguageModelsAre2020}, rewards~\cite{ouyangTrainingLanguageModels2022}, or preferences~\citep{christianoDeepReinforcementLearning2017a,rafailovDirectPreferenceOptimization2023}, enables pre-trained models to excel in downstream tasks through supervised or reinforcement learning. This recipe has given rise to models that rival human capabilities on a variety of tasks and are poised to surpass human intelligence if further scaled~\citep{ai-risk-open-letter}. However, these advances pose a pressing challenge in the continual acquisition of quality labels and the alignment of model behaviors with our desired outcomes.

\begin{figure}[t]
\centering
\vspace{-20pt}
\includegraphics[width=0.70\linewidth]{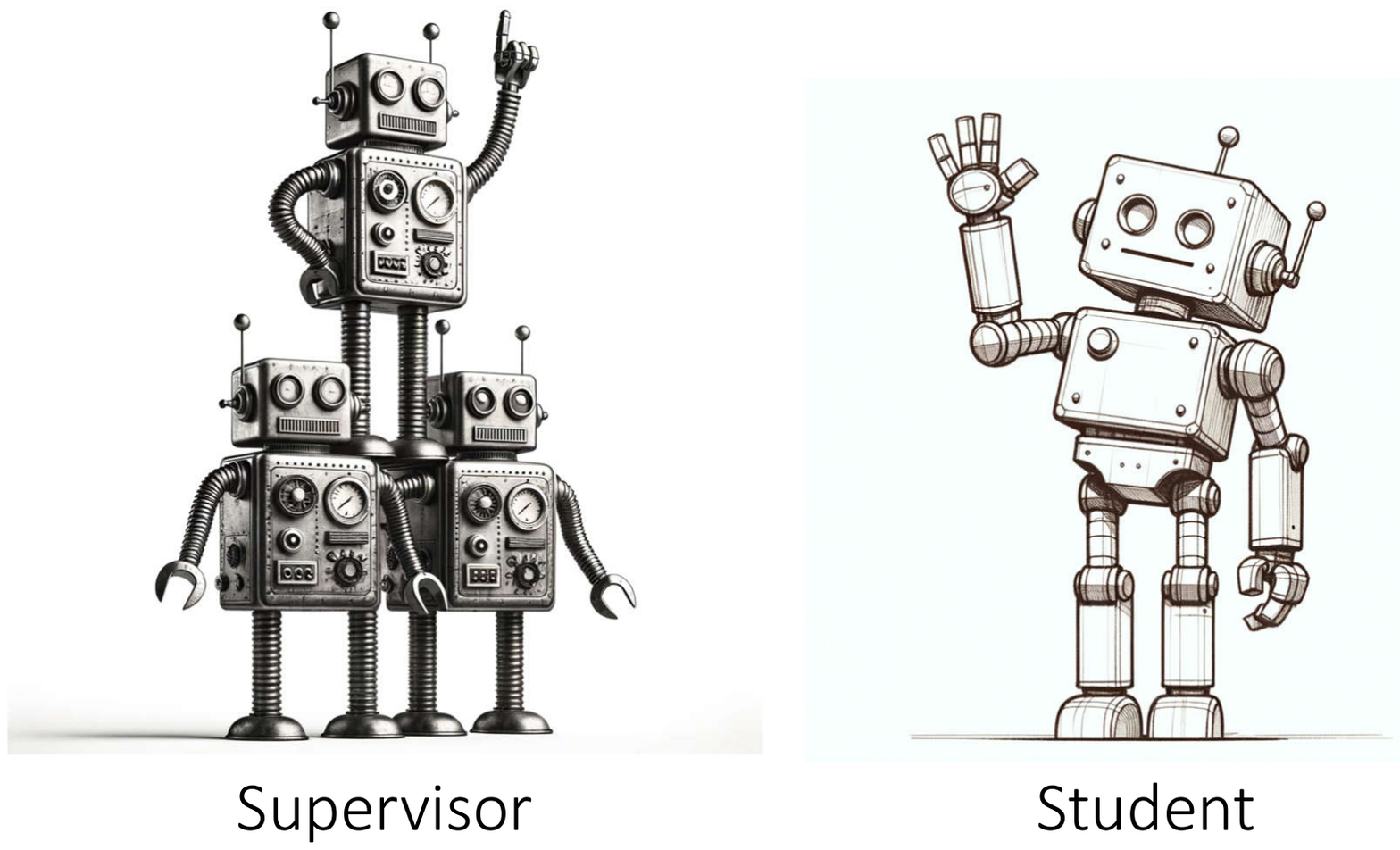}
\vspace{-25pt}
\caption{Illustration of co-supervised learning for weak-to-strong generalization. We revisit the hierarchical mixture-of-experts method in the context of superalignment, and present an approach that leverages multiple weak supervisors with different specializations to collectively supervise a strong student model. 
}
\vspace{-10pt}
\label{fig:pull}
\end{figure}

In response to this challenge, recent work has started to explore the potential of fine-tuning a strong pre-trained model under the guidance of a comparably weak supervisor~\citep{burnsWeaktoStrongGeneralizationEliciting2023}. It has been shown that, despite annotation noises, a strong student can learn from its weak supervisor and outperform it on specific tasks, such as visual recognition~\citep{dengImageNetLargescaleHierarchical2009} and language understanding~\citep{weiFinetunedLanguageModels2021}. This intriguing phenomenon brings a glimmer of hope for superalignment~\citep{superalignment}, but is still in its infancy. For instance, the effectiveness of weak-to-strong generalization tends to diminish, when the teacher's capability lags far behind the student's. Considering the inherent limits of a human supervisor against the continual growth of large pre-trained models, how can we foster more effective weak-to-strong generalization?

In this work, we propose to improve weak-to-strong generalization by incorporating multiple weak supervisors. In particular, we seek to assemble and harness a diverse set of specialized teachers, as opposed to a single generalist one, that collectively supervises a strong student. We call this {Co-Supervised Learning (CSL)}, a paradigm that conceptually mirrors the multidisciplinary nature of human educational systems and technically traces its roots back to the classical hierarchical mixture-of-experts models ~\citep{jordanHierarchicalMixturesExperts1993} and their modern variants~\citep{shazeerOutrageouslyLargeNeural2016}. Traditional efforts in mixture-of-experts focus on composing a model with multiple trainable experts~\citep{yukselTwentyYearsMixture2012,fedusReviewSparseExpert2022}. We instead argue that the experts in the alignment context are rather fixed -- they can hardly be further improved -- and yet they are already available in various sub-domains (\eg, human specialists). Individually, each may not come close to the capability of the strong student, but collectively, they can possess a wealth of knowledge relevant to the target objective, as illustrated in~\cref{fig:pull}. This premise motivates us to study co-supervised learning with two specific components.

{\it Component 1: teacher assignment.} One question at the heart of co-supervision is identifying one or a subset of teachers best suited to annotate a given unlabeled example. Classical ensemble techniques (\eg, majority voting, weighted averaging) are effective when most teachers are correct~\citep{ganaieEnsembleDeepLearning2022}. Yet, these techniques falter in challenging settings where valuable insights are confined to one or a few experts only. Some prior studies have explored the use of a gating or routing network learned atop mixture-of-experts~\citep{riquelmeScalingVisionSparse2021,zhouMixtureofExpertsExpertChoice2022}. This approach, however, hinges on the availability of ground-truth annotations, rendering it unfit for superalignment. To mitigate this, we propose to leverage the evolving competence of the student model itself. Specifically, we consider the student in its latest iteration as the closest approximation to the ground-truth annotations, and formulate co-supervision as an alternating process between student training and teacher assignment. This process is akin to an Expectation-Maximization (EM) algorithm, progressively elevating teacher specialization.

{\it Component 2: noise reduction.} Another question critical to our approach is to discard supervisory noises rather than absorb them. In the presence of a pronounced capability gap, it's almost inevitable that certain unlabeled examples may exceed the collective expertise of all weak supervisors combined. Yet, pinpointing these examples is difficult. One common strategy is to estimate noise magnitudes based on the discrepancies between the outputs of the teacher and student~\citep{jiangMentorNetLearningDataDriven2018}. Nonetheless, establishing an absolute filtering criterion remains intricate and exerts substantial influence on the outcome. In fact, prior efforts to estimate noise rates~\citep{chenUnderstandingUtilizingDeep2019,gargNoisylabelLearningSample2023} diverge far from our observations in the weak-to-strong context, due to their assumptions on the independence of models or examples. To circumvent this, we propose to leverage these interdependencies and conservatively enforce consistency between different models. This leads to two noise reduction criteria, teacher-student consistency and local-global consistency, which are simple to compute and effective in preserving a reliable subset of co-supervision.

We evaluate the proposed method on a pair of pre-trained visual representations exhibiting a significant capability gap.
Experiments on the OpenAI weak-to-strong benchmark show that our method outperforms the vanilla single-teacher baseline by over $15\%$.
We further consider another multi-domain scenario, which poses greater challenges for the prior method, yet our method yields consistent and substantial improvements. We hope our findings will furnish a co-supervised perspective for weak-to-strong generalization and serve as a stepping stone towards superalignment.

\section{Related Work}
\label{sec:related}
Our co-supervised approach to weak-to-strong generalization falls under the scope of AI alignment, with close connections to ensemble methods and learning from noisy labels.

\paragraph{AI Alignment.} The goal of AI alignment centers on steering capable models to behave in line with human values and intentions~\citep{leikeScalableAgentAlignment2018,jiAIAlignmentComprehensive2023}. This is typically pursued through fine-tuning, either imitation learning on human demonstrations~\citep{bain1995framework,atkeson1997robot} or reinforcement learning from human feedback~\citep{christianoDeepReinforcementLearning2017a,stiennon2020learning}. Both approaches, however, hinge on high-quality human supervision, a requirement that becomes increasingly challenging as pre-trained models grow in capability~\citep{amodeiConcreteProblemsAI2016}. In response to this challenge, one line of work seeks to augment the capability of human supervision. Examples include task decomposition~\citep{christiano2018supervising}, multi-agent debates~\citep{irving2018ai}, logical verification~\citep{fluriEvaluatingSuperhumanModels2023}, and many more~\citep{demski2019embedded,hubinger2019risks}, each offering a unique angle to bolster scalable oversight. Yet, how effective these methods will be in the face of superintelligence models remains unclear. Another line of research adopts a prudent and complementary stance, acknowledging the limits of human capability and aiming to supervise a strong student model with a weak teacher~\citep{superalignment,burnsWeaktoStrongGeneralizationEliciting2023}. Our work builds upon the recent attempt at the latter, exploring a new approach that leverages multiple weak teachers to collectively supervise a strong student.

\paragraph{Ensemble Methods.} Combining predictions of several models to achieve superior performance over any single one has been studied for decades in the area of ensembles~\citep{yukselTwentyYearsMixture2012,ganaieEnsembleDeepLearning2022}. One prevalent approach is to develop a collection of models, such as through bagging~\citep{breimanBaggingPredictors1996} or boosting~\citep{freundExperimentsNewBoosting1996}, and then aggregate their predictions via mechanisms like voting ~\citep{schapireBoostingMarginNew1998} or averaging~\citep{hoetingBayesianModelAveraging1999} Typically, each model in these ensembles is trained on the entire problem space, which can lead to suboptimal results when the complexity of the problem exceeds the capacity of the individual models. To mitigate this, another approach divides the problem space into distinct subspaces, each tackled by a specialized expert model~\citep{jacobsAdaptiveMixturesLocal1991a}. These expert models are usually coordinated by a routing network and are jointly trained through methods like Expectation-Maximization (EM) algorithms~\citep{jordanHierarchicalMixturesExperts1993}, variational techniques~\citep{bishopBayesianHierarchicalMixtures2002}, and differentiable top-k operations~\citep{shazeerOutrageouslyLargeNeural2016}, among others. This mixture-of-experts approach has proven effective in bolstering the collective capability without sacrificing inference efficiency~\citep{fedusReviewSparseExpert2022,jiangMixtralExperts2024}. Yet, it hinges on the presence of high-quality annotations, a prerequisite untenable for superalignment. Our work repurposes this approach to weak-to-strong generalization, framing co-supervision through the lens of mixture-of-experts.

\section{Preliminaries}
\label{sec:annot}
In this section, we will first describe the background of weak-to-strong generalization for superalignment, elucidating its challenges under large capability gaps.
We will subsequently introduce a co-supervised setting, where multiple weak supervisors with diverse specializations are available for collectively supervising the strong student model.

\subsection{Weak-to-Strong Generalization} \label{subsec:weakstrong}

\paragraph{Alignment.}
Consider an alignment problem that models the conditional probability  \( p(\vy | \vx) \), where \( \vx \) is a high-dimensional input, such as a textual prompt in language modeling or an image in vision recognition, and \( \vy \) is the corresponding output, such as a desired response or a target category. 
The model in question is typically a large neural network that has undergone extensive pre-training on a diverse range of data.
To steer its behavior towards a particular downstream task or human preference, a common practice is to fine-tune the model parameter $\vtheta$ on a labeled dataset $\mathcal{D} = \{(\vx^{(1)}, \vy^{(1)}), \ldots, (\vx^{(N)}, \vy^{(N)})\}$ by maximizing the log-likelihood of the observed data,
\begin{equation} \label{eq:finetune}
    \mathcal{L}(\boldsymbol{\theta}) =\sum_{i=1}^{N} \log p(\vy^{(i)} | \vx^{(i)}; \boldsymbol{\theta}).
\end{equation}
In an ideal scenario where the dataset is extensive in size and the labels are of high accuracy, the fine-tuned model is expected to align closely with the targeted objective.

\paragraph{Weak Supervision.}
With the emergence of superintelligent models, the capability of a human supervisor is expected to be inferior to that of the pre-trained student model. This disparity may lead to a dataset with noisy labels 
$\Tilde{\mathcal{D}} = \{(\vx^{(1)}, \Tilde{\vy}^{(1)}), \ldots, (\vx^{(N)}, \Tilde{\vy}^{(N)})\}$.
Interestingly, it has been observed that the model fine-tuned on $\Tilde{\mathcal{D}}$ can outperform its weak supervisor, a phenomenon referred to as weak-to-strong generalization~\citep{burnsWeaktoStrongGeneralizationEliciting2023}.
The efficacy of such generalization was measured through the performance gap recovered:
\begin{equation}
    {\rm PGR} := ({s - \Tilde{s}}) / ({\bar s - \Tilde{s}}),
\end{equation}
where $s$ is the performance of the model post fine-tuning, $\Tilde{s}$ is the performance of the weak supervisor, and $\bar s$ is the ceiling performance of the model if fine-tuned on the clean dataset $\mathcal{D}$.

In fact, this phenomenon has been previously studied as reversed knowledge distillation~\citep{yuanRevisitingKnowledgeDistillation2020}, prior to the rise of large pre-trained models.
However, its occurrence even with hard labels (\eg, class categories) is particularly compelling.
One hypothesis is that a large pre-trained model does not need to acquire any new capabilities through fine-tuning. Instead, it simply needs to elicit its internal knowledge relevant to the target objective.
This form of generalization opens possibilities for leveraging weaker supervision to align stronger models.

\begin{figure}[t]
\centering
\includegraphics[width=0.75\linewidth]{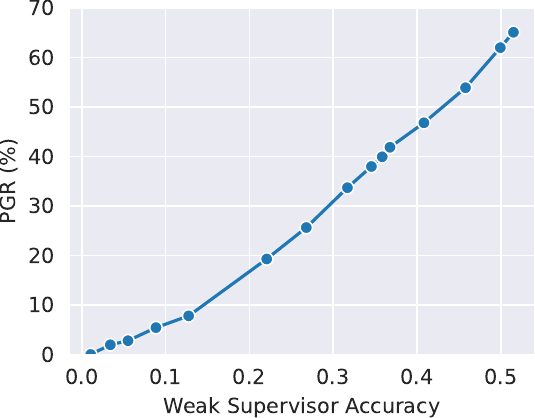}
\caption{
Effectiveness of vanilla weak-to-strong generalization.
The performance gap recovery (PGR) is notable when the performance of the supervisor is close to the ceiling performance of the strong student (0.74), but limited when the supervisor lags behind.
}
\label{fig:gap}
\end{figure}

\paragraph{Capability Gap.}
A pivotal question underpinning weak-to-strong generalization is its effectiveness in the face of superintelligent models.
While this is difficult to anticipate precisely, we seek to gain a rough estimate by examining the efficacy of this approach under varying capability gaps.
To this end, we re-train the classification head of the weak supervisor (\ie, AlexNet~\citep{krizhevskyImageNetClassificationDeep2012}) in the OpenAI weak-to-strong vision benchmark with different numbers of iterations.
\cref{fig:gap} summarizes the resulting PGR between each of these weak supervisors and the strong student model (\ie, Dino-VIT~\citep{caron2021emerging}).
It can be observed that the effectiveness of weak-to-strong generalization is tied to the capability gap.
Notably, when the performance of the weak supervisor is close to the ceiling performance of the strong student (0.74), the PGR is significant. 
However, as the capability gap widens, the PGR decreases monotonically, raising questions about the efficacy of the vanilla weak-to-strong approach for superalignment.

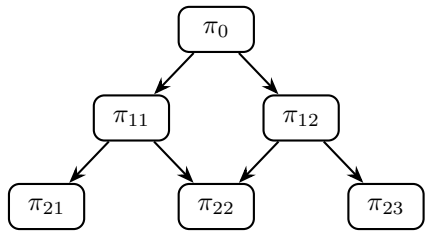
\begin{figure}[t]
\centering
\begin{tikzpicture}[
    node distance=0.55cm and 0.1cm,
    mynode/.style={draw, thick, rectangle, rounded corners, minimum width=1cm, minimum height=0.6cm, align=center},
    arrow/.style={-Stealth, thick}
]

% Nodes
\node[mynode] (pi0) {$\pi_0$};
\node[mynode] (pi11) [below left=of pi0] {$\pi_{11}$};
\node[mynode] (pi12) [below right=of pi0] {$\pi_{12}$};
\node[mynode] (pi21) [below left=of pi11] {$\pi_{21}$};
\node[mynode] (pi22) [below right=of pi11] {$\pi_{22}$};
\node[mynode] (pi23) [below right=of pi12] {$\pi_{23}$};

% Edges
\draw[arrow] (pi0) -- (pi11);
\draw[arrow] (pi0) -- (pi12);
\draw[arrow] (pi11) -- (pi21);
\draw[arrow] (pi11) -- (pi22);
\draw[arrow] (pi12) -- (pi22);
\draw[arrow] (pi12) -- (pi23);

\end{tikzpicture}
\caption{An example of two-level hierarchical weak supervisors. A generalist supervisor $\pi_0$ is first branched into two specialists $\{\pi_{11}, \pi_{12}\}$ and further branched into three $\{\pi_{21}, \pi_{22}, \pi_{23}\}$. While each specialist focuses on only a segment of the problem domain, the combined expertise at each level ensures domain coverage.}
\label{fig:hme}
\end{figure}

% each specialist focuses on a specific segment of the problem domain, their combined expertise across the hierarchy guarantees comprehensive domain coverage

\subsection{Co-Supervised Learning} \label{subsec:co-supervision}
In light of the challenge highlighted above, we explore a new setting: {\it co-supervised learning}. 
It extends the weak supervision in \cref{subsec:weakstrong} by incorporating multiple weak supervisors.
In particular, we consider a hierarchical structure of weak supervisors $\{\pi_0, \pi_1, \ldots, \pi_K\}$ across $K$ different levels, with each level hosting a set of supervisors $\pi_k = \{\pi_{k1}, \ldots, \pi_{km}\}$, distinguished by their unique areas of expertise.
This structure divides the problem domain into a series of sub-domains, aiming to allow specialized supervisors to oversee the strong students more adeptly within their sub-domains.
\cref{fig:hme} illustrates a special case of two-level specialization $\pi_{0:2}$, in which the collective supervision can be decomposed as 
\begin{equation*}
\begin{split}
    & p(\vy | \vx; \pi_{0:2}) = \sum_{\vz_1} p(\vz_1 | \vx; \pi_{0:1}) p(\vy | \vx, \vz_1; \pi_{1:2}) \\
    & = \sum_{\vz_1} p(\vz_1 | \vx; \pi_{0:1}) \sum_{\vz_2} p(\vz_2 | \vx, \vz_1; \pi_{1:2}) p(\vy | \vx, \vz_2; \pi_{2}),
\end{split}
\end{equation*}
where $\vz_k$ is a discrete variable indicating the sub-domain to which the input $\vx$ belongs at level $k$.
We omit the sample index in superscript to simplify the notation.
In general, the branching factor $m$ (number of specialized supervisors) is supposed to grow at each level.
Larger branching factors lead to higher collective capacity but at the expense of increased difficulty in estimating $\vz_k$, which is critical to the efficacy of co-supervised learning.

\section{Method}
\label{sec:method}
In this section, we will present a method to solve co-supervised learning defined in~\cref{subsec:co-supervision}. 
We will first describe a teacher assignment method to estimate the most appropriate supervisor and then introduce a noise reduction strategy to further improve its annotation quality.

\subsection{Teacher Assignment} \label{subsec:assignteacher}

One challenge in co-supervised learning stems from the unknowns of both the target output 
$\vy$ and the latent variable $\vz$.
If we had the ground-truth information about one, estimating the other would become substantially easier.
This property motivates us to explore an EM approach that alternates between expectation (teacher assignment) and maximization (student training) steps.

\begin{figure}[t]
\centering
\vspace{-20pt}
\includegraphics[width=0.725\linewidth]{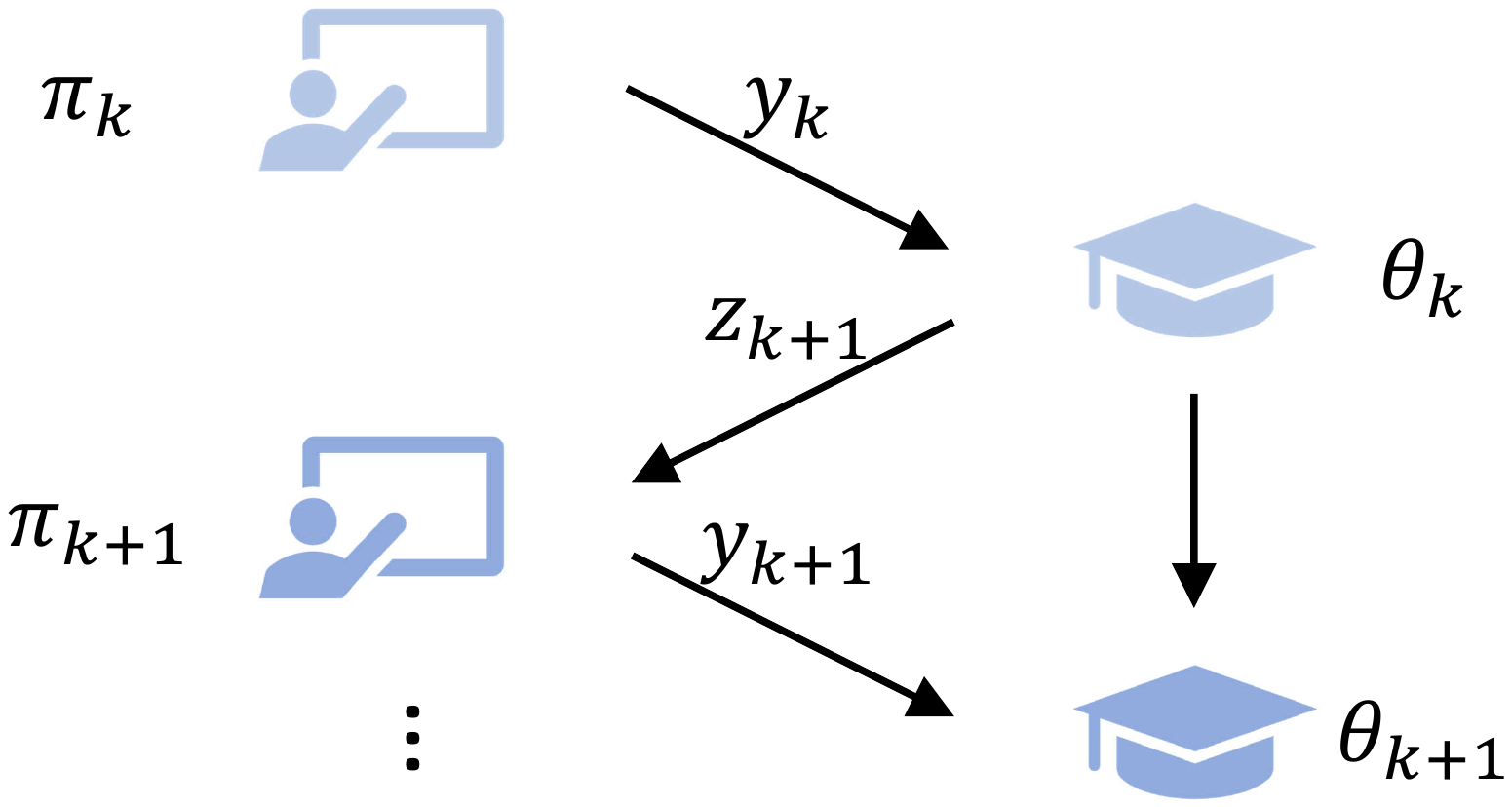}
\vspace{-26pt}
\caption{
Illustration of the alternating teacher assignment and student training processes.
The output from the latest student serves as a proxy for the target, guiding the selection of the most appropriate weak supervisor. 
The chosen supervisor is then utilized to enhance the fine-tuning of the strong student.
}
\label{fig:em}
\end{figure}

Concretely, we start the fine-tuning of the strong student $\vtheta_0$ with the generalist supervisor $\pi_0$, an initialization step equivalent to the vanilla weak-to-strong generalization~\citep{burnsWeaktoStrongGeneralizationEliciting2023}.
Next, we take the output of the student model $\hat \vy$ as an approximation of the target label and compute the expected value of the latent variable. This amounts to estimating the posterior distribution of the latent variable,
\begin{equation}
p(\vz_k | \vx, \hat \vy; \pi_{k}) = \frac{p(\vz_k | \vx; \pi_{k}) p(\hat \vy | \vx, \vz_k; \pi_{k})}{\sum_{\vz_k} {p(\vz_k | \vx; \pi_{k}) p(\hat \vy | \vx, \vz_k; \pi_{k})}}.
\end{equation}
When the specialized weak supervisors have no overlapping expertise, \ie, a hard split of the problem domain, the estimate of the latent variable reduces to find the most probable value maximizing the posterior,
\begin{equation}\label{eq:assignteacher}
\hat \vz_k := \argmax_{\vz_k} p(\vz_k | \vx; \pi_{k}) p(\hat \vy | \vx, \vz_k; \pi_{k}).
\end{equation}
Here, the first term can be modeled by a domain classifier or simplified as a uniform prior distribution.
The second term is provided by each of the weak supervisors, in favor of the one whose output is closest to that of the latest student.

While the teacher assignment process through~\cref{eq:assignteacher} is dependent on the learned student, the annotation ${\Tilde{\vy}}$ given by the assigned supervisor is not necessarily the same as $\hat \vy$ from the student.
For instance, the strong student may have learned to make an informed guess about the sub-domain of a given example but still lacks the ability to predict the precise target, where the specialized supervisor excels.
We therefore fine-tune the strong student under collective supervision at the updated level of specialization:
\begin{equation}\label{eq:trainstudent}
\vtheta_k := \argmax_{\vtheta_k} \E_{\Tilde{\mathcal{D}}} \left[ \log p(\Tilde{\vy} | \vx; \vtheta_k) \right].
\end{equation}
\cref{fig:em} illustrates the alternating process between \cref{eq:assignteacher} and \cref{eq:trainstudent}. By progressively leveling up the collective capability of the weak supervisors, our method enables the student to learn from increasingly specialized experts, thereby approaching alignment that no individual supervisors could achieve alone.

\subsection{Noise Reduction}

As described above, combining the expertise of multiple specialized supervisors is expected to boost the capacity of weak supervisors. 
However, their combined knowledge may still fall short of that possessed by a strong student pre-trained on broader data. In the presence of noisy annotations, it is generally preferable to disregard them rather than absorb the misleading information.
To identify such examples, we start with estimating the relative noise magnitudes, as suggested by prior research~\citep{jiangMentorNetLearningDataDriven2018,hanCoteachingRobustTraining2018}. 
This involves comparing the annotations provided by the weak supervisor with the prediction made by the student model,
\begin{equation}
    \epsilon := d(\vy, \Tilde{\vy}) \propto d(\hat \vy, \Tilde{\vy})
\end{equation}
where $d$ is a distance measure, \eg, cross-entropy for classification tasks or mean squared error for regression tasks.

\begin{figure}[t]
\centering
\begin{tikzpicture}[
    node distance=0.8cm and 1.2cm,
    auto,
    semithick,
    >={Stealth[scale=1.2]},
    mynode/.style={draw, thick, rectangle, rounded corners, minimum width=1.2cm, minimum height=0.8cm, align=center}, % same as your other figure
    arrow/.style={-Stealth, thick}
]

  % Nodes
  \node[mynode] (pi1) {$\pi_{k1}$};
  \node[mynode] (pi2) [below=of pi1] {$\pi_{km}$};

  \node[draw, circle, thick, minimum size=1cm, inner sep=0] (theta1) [right=of pi1] {$\theta_{k1}$};
  \node[draw, circle, thick, minimum size=1cm, inner sep=0] (theta2) [right=of pi2] {$\theta_{km}$};

  % Calculate the vertical middle for the rightmost node
  \coordinate (Middle) at ($(theta1)!0.5!(theta2)$);
  \node[draw, circle, thick, minimum size=1cm, inner sep=0] (theta) [right=of Middle] {$\theta_{k}$};

  % Lines
  \draw[arrow] (pi1) -- (theta1);
  \draw[arrow] (pi2) -- (theta2);

  \draw[arrow] (theta1) -- (theta);
  \draw[arrow] (theta2) -- (theta);

  % Dots - adjust the position manually using a slight vertical offset
  \node at ($(pi1)!0.5!(pi2) + (0,0.1)$) {\vdots};
  \node at ($(theta1)!0.5!(theta2) + (0,0.1)$) {\vdots};

\end{tikzpicture}
\caption{Progressive noise reduction via teacher-student consistency ({\color{black}{$\pi_{km} \rightarrow \theta_{km}$}}) and local-global consistency ({\color{black}{$\theta_{km} \rightarrow \theta_{k}$}}).}
\label{fig:tss}
\end{figure}
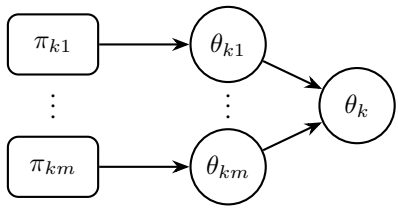

When the noise rate $\eta$ is known, prioritizing training examples by $d(\hat \vy, \Tilde{\vy})$ and removing a proportion (\eg, $1-\eta$) from the higher end often yields improved performance~\citep{jiangMentorNetLearningDataDriven2018}.
This {\em small-loss trick} has been widely adopted in cases where some training samples can be properly relabeled by expert annotators.
Oftentimes, even a tiny set of relabeled data can provide significant insights into the noise rate and its structure~\citep{hendrycksUsingTrustedData2018}.
Nonetheless, this strategy is arguably less viable in the superalignment context due to the lack of sufficiently knowledgeable supervisors.

To mitigate this challenge, we seek to once again leverage the evolving competency of the strong student in the hierarchical structure.
We start by dividing the student training into two sequential phases via locally specialized students $\vtheta_{km}$, which allows for augmenting the annotation of each specialized supervisor $\Tilde{\vy}_{km}$ with the prediction from a corresponding specialized student $\hat{\vy}_{km}$.
In each phase, we conservatively examine the agreement between the teacher and the student:
\begin{equation} \label{eq:filter}
    d(\hat{\vy}_{km}, \Tilde{\vy}_{km}) < \bar \epsilon, \quad d(\hat{\vy}_{k}, \hat{\vy}_{km}) < \bar \epsilon,
\end{equation}
where $\bar \epsilon$ is a pre-defined threshold, \eg, top-3 class agreement used in our experiments in \cref{sec:experiment}.

\cref{fig:tss} illustrates the designed noise reduction process, which assesses the annotation of each example prior to its use in training the student model. 
While only a conservative fraction of suspicious annotations is discarded each time, iterating the process progressively leads to substantial denoising effects.
The pseudo-code of our proposed method is outlined in \cref{alg:csl}.

\begin{algorithm}[t]
\label{alg:csl}
\small
\caption{Co-Supervised Learning}
\begin{algorithmic}[1] % The number tells where the line numbering should start
\setlength{\itemsep}{.2em} % Adjust the space between lines
\REQUIRE {pre-trained student $\vtheta$, fixed supervisor $\pi_0, \pi_1, \ldots, \pi_K$}
\STATE {Initialize $\vtheta_0$ under the supervision of $\pi_0$ (\autoref{eq:finetune}).}
\FOR {$k=1 \ldots K$ }
    \STATE {Collect annotations from specialized supervisors in $\pi_k$.}
    \STATE {Assign one supervisor $\vz_k$ for each example (\autoref{eq:assignteacher}).}
    \STATE {Examine annotation noise $\epsilon$ for each example (\autoref{eq:filter}).}
    \STATE {Fine-tune the student $\vtheta_k$ on the filtered dataset (\autoref{eq:trainstudent}).}
\ENDFOR
\STATE {Return $\vtheta_K$}
\end{algorithmic}
\end{algorithm}

% \section{Analysis}
% \label{sec:theory}
% \input{sections/theory}

\section{Experiments}
\label{sec:experiment}
In this section, we will present a set of experiments to answer the following questions:
\begin{enumerate}[nosep]
    \item How does domain specialization influence the collective capability of weak supervisors under a fixed capability cap?
    \item With multiple specialized supervisors available, how effectively can our method pinpoint the most appropriate supervisor?
    \item Given collective yet noisy supervision, how effective is the proposed method for reducing annotation noises and improving weak-to-strong generalization?
    \item Finally, beyond the OpenAI weak-to-strong benchmark, how does our method perform in other multi-domain scenarios?
\end{enumerate}
To this end, we adopt the evaluation protocol of prior work~\citep{burnsWeaktoStrongGeneralizationEliciting2023} and conduct experiments in visual recognition tasks on two datasets: ImageNet~\citep{dengImageNetLargescaleHierarchical2009} and DomainNet~\citep{pengMomentMatchingMultiSource2019}.
We use the encoder of AlexNet~\citep{krizhevsky2012imagenet} provided by PyTorch library~\citep{paszkePyTorchImperativeStyle2019} as the representation of the weak supervisor, and the encoder of ViT-B/14~\citep{dosovitskiyImageWorth16x162020a} extracted from DINOv2~\citep{oquabDINOv2LearningRobust2024} as the representation of the strong student.
Each model is equipped with a one-layer linear classifier on top of its pre-trained encoder. 
To examine the efficacy of weak-to-strong generalization under a large capability gap, we train the linear head of the weak supervisor on the entire training set for one epoch and subsequently evaluate our method on the validation set.
Additionally, to disentangle the influence of reversed knowledge distillation~\citep{yuanRevisitingKnowledgeDistillation2020} from weak-to-strong generalization, we experiment with two forms of annotations, both hard labels and soft logits.
Each experiment is conducted with three seeds.

\begin{figure}[t]
\centering
\small
\includegraphics[width=0.775\linewidth]{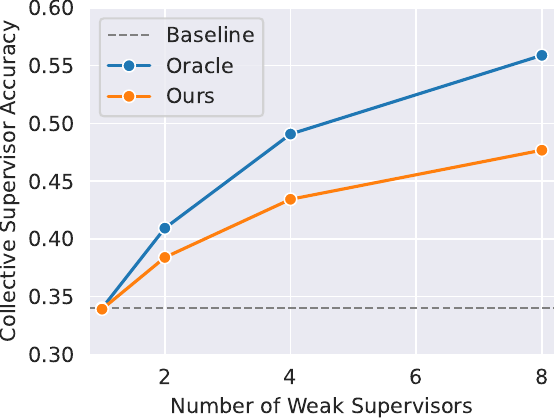}
\caption{
Collective capability of multiple specialized supervisors on ImageNet.
With each doubling of supervisor count, the test accuracy of the specialized supervisor determined by a ground-truth oracle increases by over $0.05$.
Our proposed teacher assignment method achieves approximately 2/3 of this capability boost.
}
\label{fig:assign}
\end{figure}

\subsection{Co-Supervised Learning on ImageNet} \label{exp:imagenet}

\paragraph{Setup.}
We construct specialized weak supervisors by partitioning the training set based on target classes.
Specifically, we consider three levels of specialization, with $2$, $4$, and $8$ weak supervisors overseeing $500$, $250$, and $125$ classes each, respectively.
We use $40k$ out of $50k$ datapoints from the validation set to train the linear head of the student, and evaluate its performance on the remaining $10k$ datapoints.

\paragraph{Result.}
We first examine the collective capability of multiple supervisors, comparing their test accuracy with different levels of specialization.
\cref{fig:assign} shows that the introduction of multiple specialized supervisors significantly enhances overall capability. 
Notably, our teacher assignment strategy captures a substantial portion of this capability increase. Furthermore, this enhancement does not plateau at eight supervisors, suggesting potential for further gains with the addition of more specialized supervisors.

We next assess the efficacy of our proposed noise reduction method and its impact on weak-to-strong generalization. \cref{fig:imagenet} shows that the amplified capabilities of multiple supervisors directly contribute to improvements in weak-to-strong generalization. 
Moreover, in each tested scenario, denoising weak annotations leads to consistent benefits, regardless of whether annotations are provided as hard labels or soft logits.

\begin{figure}[t]
\centering
\includegraphics[width=0.8\linewidth]{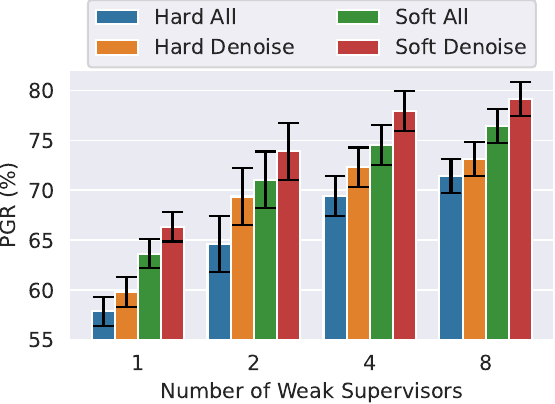}
\caption{
Effectiveness of co-supervised learning on ImageNet.
Leveraging $8$ specialized supervisors and denoising weak annotations enables our approach to outperform the baseline by more than $15\%$ in performance gap recovery with hard or soft labels.
Results are averaged across three random seeds.
}
\label{fig:imagenet}
\end{figure}

\subsection{Co-Supervised Learning on DomainNet} \label{exp:domainnet}

\paragraph{Setup.}
We construct two levels of specialized weak supervisors based on sub-domain labels from DomainNet.
At the first level, the problem domain is divided into two groups of sub-domains: {``clip", ``quick", ``sketch"} and {``info", ``paint", ``real"}.
At the second level, each supervisor is assigned to a distinct sub-domain.
We use $80k$ out of $100k$ datapoints from the validation set to train the linear head of the student, and evaluate its fine-tuning performance on the remaining $20k$ datapoints.

\paragraph{Result.}
\cref{fig:domainnet} summarize the results of our method on the DomainNet dataset. Corroborating with our observations in~\cref{exp:imagenet}, both the specialization of weak supervisors and the reduction of annotation noise contribute to more effective performance gap recovery.
It is also worth noting that the leap in performance from two to six supervisors is particularly significant with the proposed noise reduction process. This result suggests the dual advantages of incorporating multiple weak supervisors: not only does it boost the collective capacity, but it also provides richer information for more effective noise reduction.

\begin{figure}[t]
\centering
\includegraphics[width=0.8\linewidth]{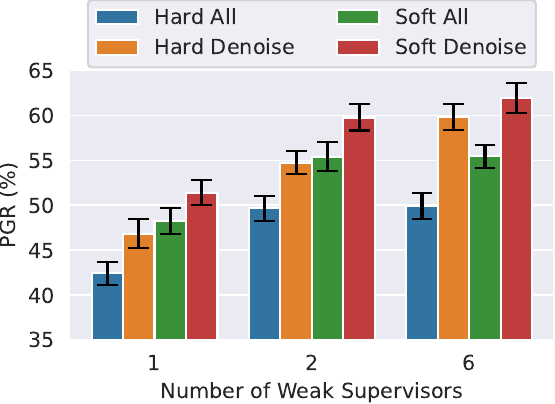}
\caption{
Effectiveness of co-supervised learning on DomainNet.
Leveraging $6$ specialized supervisors and denoising weak annotations enables our approach to outperform the baseline by $17\%$ and $14\%$ in performance gap recovery with hard and soft labels, respectively.
Results are averaged across three random seeds.
}
\label{fig:domainnet}
\end{figure}

\section{Conclusions}
\label{sec:conclusion}
\paragraph{Summary.}
We have presented a co-supervised learning approach for weak-to-strong generalization.
Leveraging a hierarchical structure of specialized supervisors, our approach significantly boosts their collective efficiency while reducing annotation noise.
The effectiveness of our method was validated through experiments on two computer vision datasets, showcasing superior performance compared to the vanilla single-supervisor baseline.

\paragraph{Limitations.} 
Our experiments were built upon the vision recognition setting due to its high computation efficiency and low resource demands. It remains unclear how our findings translate to other contexts, such as language understanding, generative modeling, and reward modeling. Understanding the efficacy and potential shortcomings of the proposed method across diverse experimental conditions might be a fruitful avenue for future research.

\section*{Acknowledgments}
This work is supported by the Swiss National Science Foundation under the Grant 2OOO21-L92326.
We thank Chelsea Finn, Parth Kothari, Dongyang Fan for helpful discussions and Riccardo Cadei, Frano Rajič, Yifan Sun for valuable feedback on early drafts.

\bibliography{bibtex/alignment, bibtex/world, bibtex/llm, bibtex/generatvie, bibtex/weakstrong, bibtex/ensemble, bibtex/noise, bibtex/dl, bibtex/domain}

\begin{thebibliography}{51}
\providecommand{\natexlab}[1]{#1}
\providecommand{\url}[1]{\texttt{#1}}
\expandafter\ifx\csname urlstyle\endcsname\relax
  \providecommand{\doi}[1]{doi: #1}\else
  \providecommand{\doi}{doi: \begingroup \urlstyle{rm}\Url}\fi

\bibitem[Amodei et~al.(2016)Amodei, Olah, Steinhardt, Christiano, Schulman, and Man{\'e}]{amodeiConcreteProblemsAI2016}
Amodei, D., Olah, C., Steinhardt, J., Christiano, P., Schulman, J., and Man{\'e}, D.
\newblock Concrete {{Problems}} in {{AI Safety}}, arXiv:1606.06565, July 2016.

\bibitem[Atkeson \& Schaal(1997)Atkeson and Schaal]{atkeson1997robot}
Atkeson, C.~G. and Schaal, S.
\newblock {Robot learning from demonstration}.
\newblock In \emph{ICML}, volume~97, pp.\  12--20. Citeseer, 1997.

\bibitem[Bain \& Sammut(1995)Bain and Sammut]{bain1995framework}
Bain, M. and Sammut, C.
\newblock {A Framework for Behavioural Cloning.}
\newblock In \emph{Machine Intelligence 15}, pp.\  103--129, 1995.

\bibitem[Bishop \& Svenskn(2002)Bishop and Svenskn]{bishopBayesianHierarchicalMixtures2002}
Bishop, C.~M. and Svenskn, M.
\newblock Bayesian hierarchical mixtures of experts.
\newblock In \emph{Proceedings of the {{Nineteenth}} Conference on {{Uncertainty}} in {{Artificial Intelligence}}}, {{UAI}}'03, pp.\  57--64, {San Francisco, CA, USA}, August 2002. {Morgan Kaufmann Publishers Inc.}
\newblock ISBN 978-0-12-705664-7.

\bibitem[Breiman(1996)]{breimanBaggingPredictors1996}
Breiman, L.
\newblock Bagging predictors.
\newblock \emph{Machine Learning}, 24\penalty0 (2):\penalty0 123--140, August 1996.

\bibitem[Brooks et~al.(2024)Brooks, Peebles, Homes, DePue, Guo, Jing, Schnurr, Taylor, Luhman, Luhman, Ng, Wang, and Ramesh]{videoworldsimulators2024}
Brooks, T., Peebles, B., Homes, C., DePue, W., Guo, Y., Jing, L., Schnurr, D., Taylor, J., Luhman, T., Luhman, E., Ng, C. W.~Y., Wang, R., and Ramesh, A.
\newblock Video generation models as world simulators.
\newblock https://openai.com/research/video-generation-models-as-world-simulators, 2024.

\bibitem[Brown et~al.(2020)Brown, Mann, Ryder, Subbiah, Kaplan, Dhariwal, Neelakantan, Shyam, Sastry, Askell, Agarwal, {Herbert-Voss}, Krueger, Henighan, Child, Ramesh, Ziegler, Wu, Winter, Hesse, Chen, Sigler, Litwin, Gray, Chess, Clark, Berner, McCandlish, Radford, Sutskever, and Amodei]{brownLanguageModelsAre2020}
Brown, T., Mann, B., Ryder, N., Subbiah, M., Kaplan, J.~D., Dhariwal, P., Neelakantan, A., Shyam, P., Sastry, G., Askell, A., Agarwal, S., {Herbert-Voss}, A., Krueger, G., Henighan, T., Child, R., Ramesh, A., Ziegler, D., Wu, J., Winter, C., Hesse, C., Chen, M., Sigler, E., Litwin, M., Gray, S., Chess, B., Clark, J., Berner, C., McCandlish, S., Radford, A., Sutskever, I., and Amodei, D.
\newblock Language {{Models}} are {{Few-Shot Learners}}.
\newblock In \emph{Advances in {{Neural Information Processing Systems}}}, volume~33, pp.\  1877--1901, 2020.

\bibitem[Burns et~al.(2023)Burns, Izmailov, Kirchner, Baker, Gao, Aschenbrenner, Chen, Ecoffet, Joglekar, Leike, Sutskever, and Wu]{burnsWeaktoStrongGeneralizationEliciting2023}
Burns, C., Izmailov, P., Kirchner, J.~H., Baker, B., Gao, L., Aschenbrenner, L., Chen, Y., Ecoffet, A., Joglekar, M., Leike, J., Sutskever, I., and Wu, J.
\newblock Weak-to-{{Strong Generalization}}: {{Eliciting Strong Capabilities With Weak Supervision}}, arXiv:2312.09390, December 2023.

\bibitem[CAIS(2023)]{ai-risk-open-letter}
CAIS.
\newblock {Statement on {AI}} risk, 2023.

\bibitem[Caron et~al.(2021)Caron, Touvron, Misra, Jégou, Mairal, Bojanowski, and Joulin]{caron2021emerging}
Caron, M., Touvron, H., Misra, I., Jégou, H., Mairal, J., Bojanowski, P., and Joulin, A.
\newblock {Emerging properties in self-supervised vision transformers}.
\newblock In \emph{Proceedings of the IEEE/CVF international conference on computer vision}, pp.\  9650--9660, 2021.

\bibitem[Chen et~al.(2019)Chen, Liao, Chen, and Zhang]{chenUnderstandingUtilizingDeep2019}
Chen, P., Liao, B.~B., Chen, G., and Zhang, S.
\newblock Understanding and {{Utilizing Deep Neural Networks Trained}} with {{Noisy Labels}}.
\newblock In \emph{Proceedings of the 36th {{International Conference}} on {{Machine Learning}}}, pp.\  1062--1070. {PMLR}, May 2019.

\bibitem[Christiano et~al.(2018)Christiano, Shlegeris, and Amodei]{christiano2018supervising}
Christiano, P., Shlegeris, B., and Amodei, D.
\newblock {Supervising strong learners by amplifying weak experts}.
\newblock \emph{arXiv preprint arXiv:1810.08575}, 2018.

\bibitem[Christiano et~al.(2017)Christiano, Leike, Brown, Martic, Legg, and Amodei]{christianoDeepReinforcementLearning2017a}
Christiano, P.~F., Leike, J., Brown, T., Martic, M., Legg, S., and Amodei, D.
\newblock Deep {{Reinforcement Learning}} from {{Human Preferences}}.
\newblock In \emph{Advances in {{Neural Information Processing Systems}}}, volume~30. {Curran Associates, Inc.}, 2017.

\bibitem[Demski \& Garrabrant(2019)Demski and Garrabrant]{demski2019embedded}
Demski, A. and Garrabrant, S.
\newblock {Embedded agency}.
\newblock \emph{arXiv preprint arXiv:1902.09469}, 2019.

\bibitem[Deng et~al.(2009)Deng, Dong, Socher, Li, and {and}]{dengImageNetLargescaleHierarchical2009}
Deng, J., Dong, W., Socher, R., Li, L., and {and}.
\newblock {{ImageNet}}: {{A}} large-scale hierarchical image database.
\newblock In \emph{2009 {{IEEE Conference}} on {{Computer Vision}} and {{Pattern Recognition}}}, pp.\  248--255, June 2009.

\bibitem[Dosovitskiy et~al.(2020)Dosovitskiy, Beyer, Kolesnikov, Weissenborn, Zhai, Unterthiner, Dehghani, Minderer, Heigold, Gelly, Uszkoreit, and Houlsby]{dosovitskiyImageWorth16x162020a}
Dosovitskiy, A., Beyer, L., Kolesnikov, A., Weissenborn, D., Zhai, X., Unterthiner, T., Dehghani, M., Minderer, M., Heigold, G., Gelly, S., Uszkoreit, J., and Houlsby, N.
\newblock An {{Image}} is {{Worth}} 16x16 {{Words}}: {{Transformers}} for {{Image Recognition}} at {{Scale}}.
\newblock In \emph{International {{Conference}} on {{Learning Representations}}}, October 2020.

\bibitem[Fedus et~al.(2022)Fedus, Dean, and Zoph]{fedusReviewSparseExpert2022}
Fedus, W., Dean, J., and Zoph, B.
\newblock A {{Review}} of {{Sparse Expert Models}} in {{Deep Learning}}, arXiv:2209.01667, September 2022.

\bibitem[Fluri et~al.(2023)Fluri, Paleka, and Tram{\`e}r]{fluriEvaluatingSuperhumanModels2023}
Fluri, L., Paleka, D., and Tram{\`e}r, F.
\newblock Evaluating {{Superhuman Models}} with {{Consistency Checks}}, arXiv:2306.09983, October 2023.

\bibitem[Freund \& Schapire(1996)Freund and Schapire]{freundExperimentsNewBoosting1996}
Freund, Y. and Schapire, R.~E.
\newblock Experiments with a new boosting algorithm.
\newblock In \emph{Proceedings of the {{Thirteenth International Conference}} on {{International Conference}} on {{Machine Learning}}}, {{ICML}}'96, pp.\  148--156, {San Francisco, CA, USA}, July 1996. {Morgan Kaufmann Publishers Inc.}
\newblock ISBN 978-1-55860-419-3.

\bibitem[Ganaie et~al.(2022)Ganaie, Hu, Malik, Tanveer, and Suganthan]{ganaieEnsembleDeepLearning2022}
Ganaie, M.~A., Hu, M., Malik, A.~K., Tanveer, M., and Suganthan, P.~N.
\newblock Ensemble deep learning: {{A}} review.
\newblock \emph{Engineering Applications of Artificial Intelligence}, 115:\penalty0 105151, October 2022.

\bibitem[Garg et~al.(2023)Garg, Nguyen, Felix, Do, and Carneiro]{gargNoisylabelLearningSample2023}
Garg, A., Nguyen, C., Felix, R., Do, T.-T., and Carneiro, G.
\newblock Noisy-label {{Learning}} with {{Sample Selection}} based on {{Noise Rate Estimate}}, arXiv:2305.19486, May 2023.

\bibitem[Han et~al.(2018)Han, Yao, Yu, Niu, Xu, Hu, Tsang, and Sugiyama]{hanCoteachingRobustTraining2018}
Han, B., Yao, Q., Yu, X., Niu, G., Xu, M., Hu, W., Tsang, I., and Sugiyama, M.
\newblock Co-teaching: {{Robust}} training of deep neural networks with extremely noisy labels.
\newblock In \emph{Advances in {{Neural Information Processing Systems}}}, volume~31. {Curran Associates, Inc.}, 2018.

\bibitem[Hendrycks et~al.(2018)Hendrycks, Mazeika, Wilson, and Gimpel]{hendrycksUsingTrustedData2018}
Hendrycks, D., Mazeika, M., Wilson, D., and Gimpel, K.
\newblock Using {{Trusted Data}} to {{Train Deep Networks}} on {{Labels Corrupted}} by {{Severe Noise}}.
\newblock In \emph{Advances in {{Neural Information Processing Systems}}}, volume~31. {Curran Associates, Inc.}, 2018.

\bibitem[Hoeting et~al.(1999)Hoeting, Madigan, Raftery, and Volinsky]{hoetingBayesianModelAveraging1999}
Hoeting, J.~A., Madigan, D., Raftery, A.~E., and Volinsky, C.~T.
\newblock Bayesian {{Model Averaging}}: {{A Tutorial}}.
\newblock \emph{Statistical Science}, 14\penalty0 (4):\penalty0 382--401, 1999.

\bibitem[Hubinger et~al.(2019)Hubinger, van Merwijk, Mikulik, Skalse, and Garrabrant]{hubinger2019risks}
Hubinger, E., van Merwijk, C., Mikulik, V., Skalse, J., and Garrabrant, S.
\newblock {Risks from learned optimization in advanced machine learning systems}.
\newblock \emph{arXiv preprint arXiv:1906.01820}, 2019.

\bibitem[Irving et~al.(2018)Irving, Christiano, and Amodei]{irving2018ai}
Irving, G., Christiano, P., and Amodei, D.
\newblock {AI safety via debate}.
\newblock \emph{arXiv preprint arXiv:1805.00899}, 2018.

\bibitem[Jacobs et~al.(1991)Jacobs, Jordan, Nowlan, and Hinton]{jacobsAdaptiveMixturesLocal1991a}
Jacobs, R.~A., Jordan, M.~I., Nowlan, S.~J., and Hinton, G.~E.
\newblock Adaptive {{Mixtures}} of {{Local Experts}}.
\newblock \emph{Neural Computation}, 3\penalty0 (1):\penalty0 79--87, March 1991.

\bibitem[Ji et~al.(2023)Ji, Qiu, Chen, Zhang, Lou, Wang, Duan, He, Zhou, Zhang, Zeng, Ng, Dai, Pan, O'Gara, Lei, Xu, Tse, Fu, McAleer, Yang, Wang, Zhu, Guo, and Gao]{jiAIAlignmentComprehensive2023}
Ji, J., Qiu, T., Chen, B., Zhang, B., Lou, H., Wang, K., Duan, Y., He, Z., Zhou, J., Zhang, Z., Zeng, F., Ng, K.~Y., Dai, J., Pan, X., O'Gara, A., Lei, Y., Xu, H., Tse, B., Fu, J., McAleer, S., Yang, Y., Wang, Y., Zhu, S.-C., Guo, Y., and Gao, W.
\newblock {{AI Alignment}}: {{A Comprehensive Survey}}, arXiv:2310.19852, November 2023.

\bibitem[Jiang et~al.(2024)Jiang, Sablayrolles, Roux, Mensch, Savary, Bamford, Chaplot, de~las Casas, Hanna, Bressand, Lengyel, Bour, Lample, Lavaud, Saulnier, Lachaux, Stock, Subramanian, Yang, Antoniak, Scao, Gervet, Lavril, Wang, Lacroix, and Sayed]{jiangMixtralExperts2024}
Jiang, A.~Q., Sablayrolles, A., Roux, A., Mensch, A., Savary, B., Bamford, C., Chaplot, D.~S., de~las Casas, D., Hanna, E.~B., Bressand, F., Lengyel, G., Bour, G., Lample, G., Lavaud, L.~R., Saulnier, L., Lachaux, M.-A., Stock, P., Subramanian, S., Yang, S., Antoniak, S., Scao, T.~L., Gervet, T., Lavril, T., Wang, T., Lacroix, T., and Sayed, W.~E.
\newblock Mixtral of {{Experts}}, arXiv:2401.04088, January 2024.

\bibitem[Jiang et~al.(2018)Jiang, Zhou, Leung, Li, and {Fei-Fei}]{jiangMentorNetLearningDataDriven2018}
Jiang, L., Zhou, Z., Leung, T., Li, L.-J., and {Fei-Fei}, L.
\newblock {{MentorNet}}: {{Learning Data-Driven Curriculum}} for {{Very Deep Neural Networks}} on {{Corrupted Labels}}.
\newblock In \emph{Proceedings of the 35th {{International Conference}} on {{Machine Learning}}}, pp.\  2304--2313. {PMLR}, July 2018.

\bibitem[Jordan \& Jacobs(1993)Jordan and Jacobs]{jordanHierarchicalMixturesExperts1993}
Jordan, M. and Jacobs, R.
\newblock Hierarchical mixtures of experts and the {{EM}} algorithm.
\newblock In \emph{Proceedings of 1993 {{International Conference}} on {{Neural Networks}} ({{IJCNN-93-Nagoya}}, {{Japan}})}, volume~2, pp.\  1339--1344 vol.2, October 1993.

\bibitem[Krizhevsky et~al.(2012{\natexlab{a}})Krizhevsky, Sutskever, and Hinton]{krizhevsky2012imagenet}
Krizhevsky, A., Sutskever, I., and Hinton, G.
\newblock {Imagenet classification with deep convolutional neural networks}.
\newblock \emph{Advances in neural information processing systems}, 25, 2012{\natexlab{a}}.

\bibitem[Krizhevsky et~al.(2012{\natexlab{b}})Krizhevsky, Sutskever, and Hinton]{krizhevskyImageNetClassificationDeep2012}
Krizhevsky, A., Sutskever, I., and Hinton, G.~E.
\newblock {{ImageNet Classification}} with {{Deep Convolutional Neural Networks}}.
\newblock In Pereira, F., Burges, C. J.~C., Bottou, L., and Weinberger, K.~Q. (eds.), \emph{Advances in {{Neural Information Processing Systems}} 25}, pp.\  1097--1105. {Curran Associates, Inc.}, 2012{\natexlab{b}}.

\bibitem[Leike \& Sutskever(2023)Leike and Sutskever]{superalignment}
Leike, J. and Sutskever, I.
\newblock {Introducing Superalignment}.
\newblock \emph{OpenAI Blog}, 2023.

\bibitem[Leike et~al.(2018)Leike, Krueger, Everitt, Martic, Maini, and Legg]{leikeScalableAgentAlignment2018}
Leike, J., Krueger, D., Everitt, T., Martic, M., Maini, V., and Legg, S.
\newblock Scalable agent alignment via reward modeling: A research direction, arXiv:1811.07871, November 2018.

\bibitem[Oquab et~al.(2024)Oquab, Darcet, Moutakanni, Vo, Szafraniec, Khalidov, Fernandez, Haziza, Massa, {El-Nouby}, Assran, Ballas, Galuba, Howes, Huang, Li, Misra, Rabbat, Sharma, Synnaeve, Xu, Jegou, Mairal, Labatut, Joulin, and Bojanowski]{oquabDINOv2LearningRobust2024}
Oquab, M., Darcet, T., Moutakanni, T., Vo, H., Szafraniec, M., Khalidov, V., Fernandez, P., Haziza, D., Massa, F., {El-Nouby}, A., Assran, M., Ballas, N., Galuba, W., Howes, R., Huang, P.-Y., Li, S.-W., Misra, I., Rabbat, M., Sharma, V., Synnaeve, G., Xu, H., Jegou, H., Mairal, J., Labatut, P., Joulin, A., and Bojanowski, P.
\newblock {{DINOv2}}: {{Learning Robust Visual Features}} without {{Supervision}}, arXiv:2304.07193, February 2024.

\bibitem[Ouyang et~al.(2022)Ouyang, Wu, Jiang, Almeida, Wainwright, Mishkin, Zhang, Agarwal, Slama, Gray, Schulman, Hilton, Kelton, Miller, Simens, Askell, Welinder, Christiano, Leike, and Lowe]{ouyangTrainingLanguageModels2022}
Ouyang, L., Wu, J., Jiang, X., Almeida, D., Wainwright, C., Mishkin, P., Zhang, C., Agarwal, S., Slama, K., Gray, A., Schulman, J., Hilton, J., Kelton, F., Miller, L., Simens, M., Askell, A., Welinder, P., Christiano, P., Leike, J., and Lowe, R.
\newblock Training language models to follow instructions with human feedback.
\newblock In \emph{Advances in {{Neural Information Processing Systems}}}, October 2022.

\bibitem[Paszke et~al.(2019)Paszke, Gross, Massa, Lerer, Bradbury, Chanan, Killeen, Lin, Gimelshein, Antiga, Desmaison, Kopf, Yang, DeVito, Raison, Tejani, Chilamkurthy, Steiner, Fang, Bai, and Chintala]{paszkePyTorchImperativeStyle2019}
Paszke, A., Gross, S., Massa, F., Lerer, A., Bradbury, J., Chanan, G., Killeen, T., Lin, Z., Gimelshein, N., Antiga, L., Desmaison, A., Kopf, A., Yang, E., DeVito, Z., Raison, M., Tejani, A., Chilamkurthy, S., Steiner, B., Fang, L., Bai, J., and Chintala, S.
\newblock {{PyTorch}}: {{An Imperative Style}}, {{High-Performance Deep Learning Library}}.
\newblock In \emph{Advances in {{Neural Information Processing Systems}}}, volume~32. {Curran Associates, Inc.}, 2019.

\bibitem[Peng et~al.(2019)Peng, Bai, Xia, Huang, Saenko, and Wang]{pengMomentMatchingMultiSource2019}
Peng, X., Bai, Q., Xia, X., Huang, Z., Saenko, K., and Wang, B.
\newblock Moment {{Matching}} for {{Multi-Source Domain Adaptation}}.
\newblock In \emph{2019 {{IEEE}}/{{CVF International Conference}} on {{Computer Vision}} ({{ICCV}})}, pp.\  1406--1415. {IEEE Computer Society}, October 2019.
\newblock ISBN 978-1-72814-803-8.

\bibitem[Rafailov et~al.(2023)Rafailov, Sharma, Mitchell, Manning, Ermon, and Finn]{rafailovDirectPreferenceOptimization2023}
Rafailov, R., Sharma, A., Mitchell, E., Manning, C.~D., Ermon, S., and Finn, C.
\newblock Direct {{Preference Optimization}}: {{Your Language Model}} is {{Secretly}} a {{Reward Model}}.
\newblock In \emph{Thirty-Seventh {{Conference}} on {{Neural Information Processing Systems}}}, November 2023.

\bibitem[Ramesh et~al.(2021)Ramesh, Pavlov, Goh, Gray, Voss, Radford, Chen, and Sutskever]{rameshZeroShotTexttoImageGeneration2021a}
Ramesh, A., Pavlov, M., Goh, G., Gray, S., Voss, C., Radford, A., Chen, M., and Sutskever, I.
\newblock Zero-{{Shot Text-to-Image Generation}}.
\newblock In \emph{Proceedings of the 38th {{International Conference}} on {{Machine Learning}}}, pp.\  8821--8831. {PMLR}, July 2021.

\bibitem[Riquelme et~al.(2021)Riquelme, Puigcerver, Mustafa, Neumann, Jenatton, Susano~Pinto, Keysers, and Houlsby]{riquelmeScalingVisionSparse2021}
Riquelme, C., Puigcerver, J., Mustafa, B., Neumann, M., Jenatton, R., Susano~Pinto, A., Keysers, D., and Houlsby, N.
\newblock Scaling {{Vision}} with {{Sparse Mixture}} of {{Experts}}.
\newblock In \emph{Advances in {{Neural Information Processing Systems}}}, volume~34, pp.\  8583--8595. {Curran Associates, Inc.}, 2021.

\bibitem[Rombach et~al.(2022)Rombach, Blattmann, Lorenz, Esser, and Ommer]{rombachHighResolutionImageSynthesis2022}
Rombach, R., Blattmann, A., Lorenz, D., Esser, P., and Ommer, B.
\newblock High-{{Resolution Image Synthesis With Latent Diffusion Models}}.
\newblock In \emph{Proceedings of the {{IEEE}}/{{CVF Conference}} on {{Computer Vision}} and {{Pattern Recognition}}}, pp.\  10684--10695, 2022.

\bibitem[Schapire et~al.(1998)Schapire, Freund, Bartlett, and Lee]{schapireBoostingMarginNew1998}
Schapire, R.~E., Freund, Y., Bartlett, P., and Lee, W.~S.
\newblock {Boosting the margin: A new explanation for the effectiveness of voting methods}.
\newblock \emph{Annals of Statistics}, 26\penalty0 (5):\penalty0 1651--1686, October 1998.

\bibitem[Shazeer et~al.(2016)Shazeer, Mirhoseini, Maziarz, Davis, Le, Hinton, and Dean]{shazeerOutrageouslyLargeNeural2016}
Shazeer, N., Mirhoseini, A., Maziarz, K., Davis, A., Le, Q., Hinton, G., and Dean, J.
\newblock Outrageously {{Large Neural Networks}}: {{The Sparsely-Gated Mixture-of-Experts Layer}}.
\newblock In \emph{International {{Conference}} on {{Learning Representations}}}, November 2016.

\bibitem[Stiennon et~al.(2020)Stiennon, Ouyang, Wu, Ziegler, Lowe, Voss, Radford, Amodei, and Christiano]{stiennon2020learning}
Stiennon, N., Ouyang, L., Wu, J., Ziegler, D., Lowe, R., Voss, C., Radford, A., Amodei, D., and Christiano, P.
\newblock {Learning to summarize with human feedback}.
\newblock \emph{Advances in Neural Information Processing Systems}, 33:\penalty0 3008--3021, 2020.

\bibitem[Touvron et~al.(2023)Touvron, Martin, Stone, Albert, Almahairi, Babaei, Bashlykov, Batra, Bhargava, Bhosale, Bikel, Blecher, Ferrer, Chen, Cucurull, Esiobu, Fernandes, Fu, Fu, Fuller, Gao, Goswami, Goyal, Hartshorn, Hosseini, Hou, Inan, Kardas, Kerkez, Khabsa, Kloumann, Korenev, Koura, Lachaux, Lavril, Lee, Liskovich, Lu, Mao, Martinet, Mihaylov, Mishra, Molybog, Nie, Poulton, Reizenstein, Rungta, Saladi, Schelten, Silva, Smith, Subramanian, Tan, Tang, Taylor, Williams, Kuan, Xu, Yan, Zarov, Zhang, Fan, Kambadur, Narang, Rodriguez, Stojnic, Edunov, and Scialom]{touvronLlamaOpenFoundation2023}
Touvron, H., Martin, L., Stone, K., Albert, P., Almahairi, A., Babaei, Y., Bashlykov, N., Batra, S., Bhargava, P., Bhosale, S., Bikel, D., Blecher, L., Ferrer, C.~C., Chen, M., Cucurull, G., Esiobu, D., Fernandes, J., Fu, J., Fu, W., Fuller, B., Gao, C., Goswami, V., Goyal, N., Hartshorn, A., Hosseini, S., Hou, R., Inan, H., Kardas, M., Kerkez, V., Khabsa, M., Kloumann, I., Korenev, A., Koura, P.~S., Lachaux, M.-A., Lavril, T., Lee, J., Liskovich, D., Lu, Y., Mao, Y., Martinet, X., Mihaylov, T., Mishra, P., Molybog, I., Nie, Y., Poulton, A., Reizenstein, J., Rungta, R., Saladi, K., Schelten, A., Silva, R., Smith, E.~M., Subramanian, R., Tan, X.~E., Tang, B., Taylor, R., Williams, A., Kuan, J.~X., Xu, P., Yan, Z., Zarov, I., Zhang, Y., Fan, A., Kambadur, M., Narang, S., Rodriguez, A., Stojnic, R., Edunov, S., and Scialom, T.
\newblock Llama 2: {{Open Foundation}} and {{Fine-Tuned Chat Models}}, arXiv:2307.09288, July 2023.

\bibitem[Wei et~al.(2021)Wei, Bosma, Zhao, Guu, Yu, Lester, Du, Dai, and Le]{weiFinetunedLanguageModels2021}
Wei, J., Bosma, M., Zhao, V., Guu, K., Yu, A.~W., Lester, B., Du, N., Dai, A.~M., and Le, Q.~V.
\newblock Finetuned {{Language Models}} are {{Zero-Shot Learners}}.
\newblock In \emph{International {{Conference}} on {{Learning Representations}}}, October 2021.

\bibitem[Yuan et~al.(2020)Yuan, Tay, Li, Wang, and Feng]{yuanRevisitingKnowledgeDistillation2020}
Yuan, L., Tay, F.~E., Li, G., Wang, T., and Feng, J.
\newblock Revisiting {{Knowledge Distillation}} via {{Label Smoothing Regularization}}.
\newblock In \emph{2020 {{IEEE}}/{{CVF Conference}} on {{Computer Vision}} and {{Pattern Recognition}} ({{CVPR}})}, pp.\  3902--3910, {Seattle, WA, USA}, June 2020. {IEEE}.
\newblock ISBN 978-1-72817-168-5.

\bibitem[Yuksel et~al.(2012)Yuksel, Wilson, and Gader]{yukselTwentyYearsMixture2012}
Yuksel, S.~E., Wilson, J.~N., and Gader, P.~D.
\newblock Twenty {{Years}} of {{Mixture}} of {{Experts}}.
\newblock \emph{IEEE Transactions on Neural Networks and Learning Systems}, 23\penalty0 (8):\penalty0 1177--1193, August 2012.

\bibitem[Zhou et~al.(2022)Zhou, Lei, Liu, Du, Huang, Zhao, Dai, Chen, Le, and Laudon]{zhouMixtureofExpertsExpertChoice2022}
Zhou, Y., Lei, T., Liu, H., Du, N., Huang, Y., Zhao, V., Dai, A.~M., Chen, Z., Le, Q.~V., and Laudon, J.
\newblock Mixture-of-{{Experts}} with {{Expert Choice Routing}}.
\newblock \emph{Advances in Neural Information Processing Systems}, 35:\penalty0 7103--7114, December 2022.

\end{thebibliography}
\bibliographystyle{icml2024}

\clearpage
%%%%%%%%%%%%%%%%%%%%%%%%%%%%%%%%%%%%%%%%%%%%%%%%%%%%%%%%%%%%%%%%%%%%%%%%%%%%%%%
%%%%%%%%%%%%%%%%%%%%%%%%%%%%%%%%%%%%%%%%%%%%%%%%%%%%%%%%%%%%%%%%%%%%%%%%%%%%%%%
% APPENDIX
%%%%%%%%%%%%%%%%%%%%%%%%%%%%%%%%%%%%%%%%%%%%%%%%%%%%%%%%%%%%%%%%%%%%%%%%%%%%%%%
%%%%%%%%%%%%%%%%%%%%%%%%%%%%%%%%%%%%%%%%%%%%%%%%%%%%%%%%%%%%%%%%%%%%%%%%%%%%%%%
\newpage
\appendix
\onecolumn

%%%%%%%%%%%%%%%%%%%%%%%%%%%%%%%%%%%%%%%%%%%%%%%%%%%%%%%%%%%%%%%%%%%%%%%%%%%%%%%
%%%%%%%%%%%%%%%%%%%%%%%%%%%%%%%%%%%%%%%%%%%%%%%%%%%%%%%%%%%%%%%%%%%%%%%%%%%%%%%

\end{document}